\documentclass[10pt,twocolumn,letterpaper]{article}

\usepackage[pagenumbers]{wacv}              %
\usepackage{multirow}
\usepackage[accsupp]{axessibility}

\definecolor{wacvblue}{rgb}{0.21,0.49,0.74}
\usepackage[pagebackref,breaklinks,colorlinks,allcolors=wacvblue]{hyperref}

\def\wacvPaperID{2782} %
\def\confName{WACV}
\def\confYear{2026}

\def\gorillazoo{Gorilla-Zoo-Berlin}
\def\gorillawild{Gorilla-SPAC-Wild}
\def\gorillatracking{Gorilla-SPAC-MoT}

\title{GorillaWatch: An Automated System for In-the-Wild Gorilla Re-Identification and Population Monitoring}

\author{
Maximilian Schall\textsuperscript{1},
Felix Leonard Knöfel\textsuperscript{1},
Noah Elias König\textsuperscript{1},
Jan Jonas Kubeler\textsuperscript{1},
Maximilian von Klinski\textsuperscript{1}, \\
Joan Wilhelm Linnemann\textsuperscript{1},
Xiaoshi Liu\textsuperscript{1},
Iven Jelle Schlegelmilch\textsuperscript{1},
Ole Woyciniuk\textsuperscript{1},
Alexandra Schild\textsuperscript{1}, \\
Dante Wasmuht\textsuperscript{2},
Magdalena Bermejo Espinet\textsuperscript{3},
German Illera Basas\textsuperscript{3},
Gerard de Melo\textsuperscript{1}
\\
\textsuperscript{1}Hasso Plattner Institute, Germany \\
\textsuperscript{2}Conservation X Labs \\
\textsuperscript{3}Sabine Plattner African Charities %
\\
{\tt\small maximilian.schall@hpi.de} %
\vspace{2mm} \\ %
\href{https://gorilla-watch.github.io}{\color[RGB]{34,139,34}Gorilla-Watch.GitHub.io}
}

\begin{document}
\maketitle
\begin{abstract}
Monitoring critically endangered western lowland gorillas is currently hampered by the immense manual effort required to re-identify individuals from vast archives of camera trap footage. The primary obstacle to automating this process has been the lack of large-scale, ``in-the-wild'' video datasets suitable for training robust deep learning models. To address this gap, we introduce a comprehensive benchmark with three novel datasets: \gorillawild{}, the largest video dataset for wild primate re-identification to date; \gorillazoo{}, for assessing cross-domain re-identification generalization; and \gorillatracking{}, for evaluating multi-object tracking in camera trap footage.

Building on these datasets, we present \textbf{GorillaWatch}, an end-to-end pipeline integrating detection, tracking, and re-identification. To exploit temporal information, we introduce a multi-frame self-supervised pretraining strategy that leverages consistency in tracklets to learn domain-specific features without manual labels. To ensure scientific validity, a differentiable adaptation of AttnLRP verifies that our model relies on discriminative biometric traits rather than background correlations. Extensive benchmarking subsequently demonstrates that aggregating features from large-scale image backbones outperforms specialized video architectures. Finally, we address population counting without identity labels by integrating spatiotemporal and social-group constraints into standard clustering, which mitigates over-segmentation on in-the-wild footage. We publicly release all code and datasets to facilitate scalable, non-invasive monitoring of endangered species.  

\end{abstract}

\section{Introduction}
Western Lowland Gorillas (\textit{Gorilla gorilla gorilla}), our closest living relatives after chimpanzees and bonobos, are critically endangered~\cite{maisels_f_gorilla_2018}. Camera traps have emerged as the principal non-invasive solution for wildlife monitoring, generating vast archives of video data that capture natural behavior and individual identities~\cite{beery_recognition_2018}. However, the sheer volume of footage creates a critical bottleneck; conservation efforts are currently hindered by the immense manual effort required to analyze these videos, making rapid population assessment nearly impossible.

The motivation for large-scale monitoring is twofold. In conservation, accurate demographic models are essential for assessing population viability, tracking disease transmission (e.g., Ebola~\cite{bermejo_ebola_2006}), and monitoring injuries from human-wildlife conflict. In the scientific domain, long-term observation of social dynamics and behaviors provides data for understanding the evolutionary foundations of human sociality~\cite{morrison_hierarchical_2019}. Both goals rely on the ability to reliably track distinct individuals over extended periods, a task that presents considerable logistical and analytical challenges.

Automating this process requires robust re-identification (re-ID) pipelines, yet developing them for wildlife comes with difficulties not present in human surveillance~\cite{otarashvili_multispecies_2024}. The primary obstacle has been the lack of large-scale, annotated video datasets captured ``in the wild.'' Existing primate datasets~\cite{brookes_dataset_2021,freytag_chimpanzee_2016} are often limited in scale, rely on static images, or feature captive animals, failing to prepare models for the challenging lighting, occlusion, and pose variations encountered in the rainforest. 

To address these gaps, we propose a comprehensive benchmark suite of three novel datasets. \gorillawild{} is the largest in-the-wild video dataset for primate re-identification to date, specifically curated to evaluate the challenging cross-encounter setting. To test robustness and generalizability, we introduce \gorillazoo{}, a cross-domain dataset from a controlled environment, and \gorillatracking{}, a meticulously annotated benchmark for multi-object tracking.

\begin{figure*}[!ht]
    \centering
    \includegraphics[width=\linewidth]{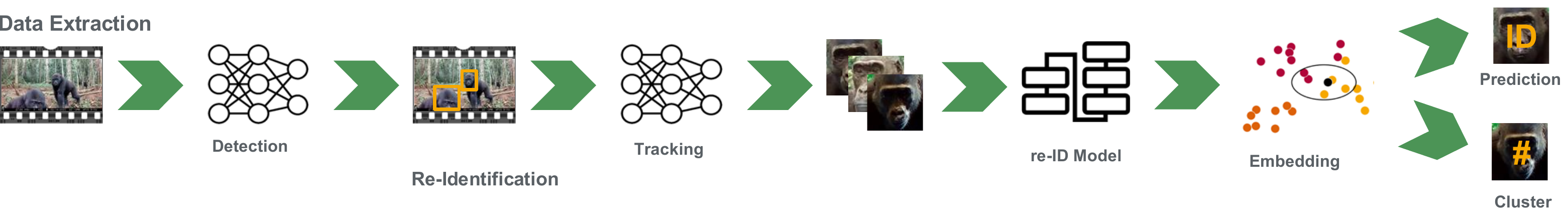}
    \caption{Overview of our complete re-identification pipeline.}
    \label{fig:pipeline}
    \vspace{-2mm}
\end{figure*}

Building on these resources, we present \textbf{GorillaWatch}, a complete end-to-end pipeline that integrates detection, tracking, and re-identification. It supports two primary workflows: retrieval-based tracking for known individuals and constrained clustering for estimating the number of distinct individuals in footage without identity labels. An illustrative overview is given in Figure~\ref{fig:pipeline}. Our work makes the following technical contributions:
\begin{enumerate}
    \item We introduce a \textbf{multi-frame self-supervised pretraining strategy} that adapts DINOv2~\cite{oquab_dinov2_2023} to leverage temporal consistency in tracklets, learning powerful domain-specific features without manual labels.
    \item We introduce a \textbf{differentiable adaptation of Attn\-LRP}~\cite{achtibat_attnlrp_2024} to interpret embedding-based retrieval. This allows us to verify that our model relies on valid biometric traits (e.g., facial features) rather than background correlations, ensuring scientific validity.
    \item We perform a \textbf{systematic evaluation of architectures}, comparing video transformers against ensemble-based image foundation models. We demonstrate that aggregating features from large-scale image backbones outperforms specialized video models in this data-scarce domain.
    \item We address the problem of population counting without identity labels by \textbf{integrating spatiotemporal constraints} into standard clustering algorithms. We show that off-the-shelf clustering leads to severe over-segmentation on in-the-wild footage, and that applying cannot-link constraints (e.g., for simultaneous detections or known social-group membership) brings the population estimates of HAC and DBSCAN close to the true population size without requiring novel clustering approaches.

\end{enumerate}

\section{Related Work}

\subsection{Re-Identification and Foundation Models}
Individual re-identification has transitioned from methods based on handcrafted features of distinctive patterns \cite{hiby_tiger_2009, bolger_computerassisted_2012, Crall2013HotSpotterP} to deep learning architectures. Early convolutional neural networks such as DeepReID \cite{li_deepreid_2014} and FaceNet \cite{schroff_facenet_2015} established the use of metric learning with triplet loss \cite{Hermans2017InDO}, while later work introduced angular margin frameworks to improve feature discrimination \cite{liu_sphereface_2017}. While these methods were developed for human re-ID, the field of animal re-ID has adapted them for species-specific applications \cite{freytag_chimpanzee_2016, brust_towards_2017, Brookes2022EvaluatingCE, adam_seaturtleid2022_2024, laskowski2023gorillavision} and, more recently, for multi-species contexts using transformers \cite{he_transreid_2021, otarashvili_multispecies_2024, cermak_wildlifedatasets_2023}. The advent of foundation models \cite{bommasani_opportunities_2022}, particularly self-supervised vision transformers like DINO \cite{caron_emerging_2021} and DINOv2 \cite{oquab_dinov2_2023}, has revolutionized feature extraction, providing powerful, general-purpose representations that can be fine-tuned for specialized domains like animal re-ID.

\subsection{Video Models and Temporal Processing}
To leverage temporal information, video transformers extend spatial attention mechanisms across the time dimension. Seminal works like TimeSformer \cite{bertasius_is_2021-1} and ViViT \cite{arnab_vivit_2021} established key architectures, often by adapting powerful pretrained image models that require full fine-tuning. More recent approaches include large-scale, pretrained video foundation models like InternVideo2 \cite{wang_internvideo2_2024}, which achieve state-of-the-art results. A parallel, more computationally efficient trend is parameter-efficient adaptation, where the pretrained image model is frozen. Methods like AIM \cite{yang_aim_2023} exemplify this by training only lightweight adapters to add temporal reasoning. However, a majority of these models are optimized for action recognition, a task focused on semantically classifying short clips. Their adaptation for re-ID remains under-explored and is a central focus of our work.

\subsection{Multimodal Fusion and Temporal Aggregation}
Multimodal approaches combine complementary features for robust identification. ShARc~\cite{zhu_sharc_2023} integrates shape and appearance through attention-based aggregation for human re-ID, while approaches like MaskReID~\cite{qi_maskreid_2019} leverage segmentation masks. Temporal aggregation methods range from basic pooling to sophisticated attention mechanisms~\cite{fu_sta_2019, wang_pyramid_2021, eom_video-based_2021}. Animal-specific work includes ReckonGait~\cite{min_recongait_2024} using spatio-temporal CNN features, and recent video-based approaches like PolarBearVidID~\cite{Zuerl2023PolarBearVidIDAV}. Cross-species approaches~\cite{li_adaptive_2024} demonstrate the potential for generalized animal re-ID frameworks.

\subsection{Detection and Tracking}
An effective re-ID system relies on a robust upstream pipeline for detection and multi-object tracking (MOT). Lightweight and efficient object detectors like the YOLO family \cite{Terven2023ACR, Varghese2024YOLOv8AN} are now standard. MOT algorithms have progressed from early methods like SORT \cite{Bewley2016SimpleOA}, which used motion cues, to modern approaches that incorporate appearance information. DeepSORT \cite{Wojke2017SimpleOA} added a re-ID model for better association, while recent trackers like ByteTrack \cite{Zhang2021ByteTrackMT}, BoT-SORT \cite{Aharon2022BoTSORTRA}, and BoostTrack \cite{Stanojevic2024BoostTrackUT} have set new benchmarks \cite{Milan2016MOT16AB, Dendorfer2020MOT20AB} by skillfully managing low-confidence detections and combining motion and appearance cues. Our work integrates and evaluates these state-of-the-art trackers to generate the high-quality tracklets that are essential for accurate downstream re-ID.

\subsection{Wildlife Datasets and Benchmarks}
\label{sec:rel_work_datasets}
Existing animal re-identification datasets suffer from several limitations that restrict 
their real-world applicability. Many focus on still images~\cite{cermak_wildlifedatasets_2023, li_atrw_2020, freytag_chimpanzee_2016, nepovinnykh_sealid_2022, adam_seaturtleid2022_2024, blount_flukebook_2022, adam_wildlifereid-10k_2025} 
instead of video, are captured in controlled environments like zoos or farms~\cite{brookes_dataset_2021, bergamini_multi-views_2018, odo_re-identification_2025, Zuerl2023PolarBearVidIDAV}, 
or adopt closed-set evaluation protocols unsuitable for wild populations where new individuals 
frequently appear~\cite{brust_towards_2017}. While recent large-scale camera trap benchmarks 
have focused on multi-animal tracking~\cite{wasmuht_sa-fari_2025, Zhang2022AnimalTrackAB, Han2024Multianimal3S, Jiang2025SAM2MOTAN} 
and behavior recognition~\cite{brookes_panaf20k_2024}, they primarily address intra-video coherence.

\section{The Gorilla Datasets}
\begin{figure}[ht]
  \centering

  \begin{subfigure}[t]{0.19\linewidth}
    \includegraphics[width=\linewidth]{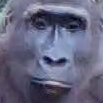}
  \end{subfigure}\hfill
  \begin{subfigure}[t]{0.19\linewidth}
    \includegraphics[width=\linewidth]{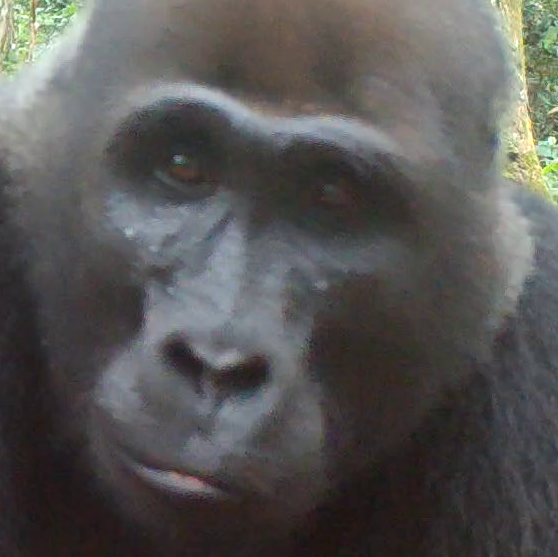}
  \end{subfigure}\hfill
  \begin{subfigure}[t]{0.19\linewidth}
    \includegraphics[width=\linewidth]{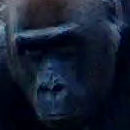}
  \end{subfigure}\hfill
  \begin{subfigure}[t]{0.19\linewidth}
    \includegraphics[width=\linewidth]{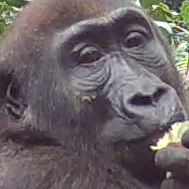}
  \end{subfigure}\hfill
  \begin{subfigure}[t]{0.19\linewidth}
    \includegraphics[width=\linewidth]{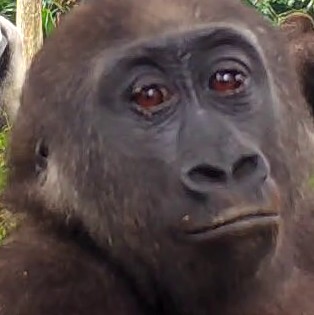}
  \end{subfigure}

  \vspace{0.1em}
  \begin{subfigure}[t]{0.19\linewidth}
    \includegraphics[width=\linewidth]{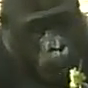}
  \end{subfigure}\hfill
  \begin{subfigure}[t]{0.19\linewidth}
    \includegraphics[width=\linewidth]{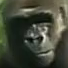}
  \end{subfigure}\hfill
  \begin{subfigure}[t]{0.19\linewidth}
    \includegraphics[width=\linewidth]{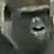}
  \end{subfigure}\hfill
  \begin{subfigure}[t]{0.19\linewidth}
    \includegraphics[width=\linewidth]{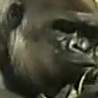}
  \end{subfigure}\hfill
  \begin{subfigure}[t]{0.19\linewidth}
    \includegraphics[width=\linewidth]{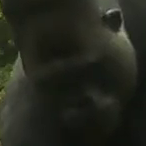}
  \end{subfigure}

  \caption{Sample images from the \gorillawild{} dataset (top row) and the \gorillazoo{} dataset (bottom row).}
  \label{fig:image_examples}
  \vspace{-2mm}
\end{figure}

To bridge this gap, we introduce three complementary datasets designed to enable robust, open-world video re-identification. \textbf{\gorillawild{}} provides in-the-wild, cross-encounter video data for training robust re-identification models; \textbf{\gorillazoo{}} is designed for cross-domain evaluation; and the meticulously annotated \textbf{\gorillatracking{}} dataset is introduced to benchmark multi-object tracking performance. Sample images from our datasets are shown in Figure~\ref{fig:image_examples}.

\subsection{\gorillawild{}}
\gorillawild{} is a image and video dataset containing 160,818 samples extracted from 1,371 tracklets across 738 camera trap videos. The footage captures 135 distinct Western Lowland Gorillas in Odzala-Kokoua National Park, Republic of Congo, recorded between October 2018 and March 2023. 
Camera traps were deployed at 33 locations specifically chosen where evidence of non-seasonal root feeding behavior had been observed, primarily surrounding \textit{Maranthes glabra} trees. This targeted placement provides a unique vantage point for capturing natural foraging behaviors and social interactions. Ground truth identity labels were provided by primate researchers with over 15 years of field experience with this specific population.

Images are extracted from automatically generated tracklets using YOLOv8\textsubscript{Nano}~\cite{Varghese2024YOLOv8AN} and BoostTrack++~\cite{Stanojevic2024BoostTrackUT}. Alongside each face, we also extract the corresponding full-body crop, creating a paired dataset that enables face-body analysis. To ensure quality, we filter for faces that are at least 50×50 pixels.

Distinct from previous datasets, \gorillawild{} is explicitly structured around ``encounters,'' defined as all recordings of a group at a specific camera location within a single calendar day. This structure is critical for realistic evaluation. Ideally, queries and the database would be split into distinct, time-separated sets; however, given the sporadic nature of wildlife sightings, such a split would result in a prohibitively small evaluation set with few positive matches. Consequently, we adopt a \textbf{cross-encounter protocol}, where queries must match identities in a different encounter. This approach maximizes data utilization while ensuring models are tested on their ability to handle significant environmental and pose variations, rather than trivial frame-to-frame matching.

The dataset is divided into four identity-disjoint partitions. We explicitly separate individuals with only a single recorded encounter (27 individuals) to serve as challenging gallery distractors. The remaining individuals, with multiple encounters, are split into 70\% Train (75 individuals), 15\% Validation (17 individuals), and 15\% Test (16 individuals).

\subsection{\gorillazoo{}}
The \gorillazoo{} dataset serves as a cross-domain evaluation benchmark, comprising 153 videos of 5 individual western lowland gorillas using 3 distinct cameras. The videos were recorded over a span of 3 months at the Berlin Zoo. It contains 188,679 annotated face bounding boxes along the corresponding body crops, across 275 tracklets, providing a controlled environment distinct from the wild camera trap conditions of \gorillawild{}. Due to camera placement and enclosure constraints, face detections are smaller than in the wild dataset. The preprocessing is the same as for \gorillawild{}. 

This dataset introduces domain shifts distinct from the rainforest setting, including artificial structures (e.g., buildings, glass), different lighting conditions, and different camera angles. Crucially, the individuals in \gorillazoo{} do not overlap with \gorillawild{}, and the camera hardware differs from the cameras used in the field. This allows us to test a model's ability to generalize its learned features to a completely new setting. Preprocessing follows the same pipeline as for \gorillawild{}.

\subsection{\gorillatracking{}}

To evaluate multi-object tracking, we introduce \gorillatracking{}, a dataset of 25 videos with dense, manual frame-by-frame tracking annotations. We categorize the dataset by interaction density. The 16 ``Hard'' videos feature frequent interactions between multiple individuals (mean $\sim$4), creating scenarios prone to identity switches due to occlusion. The remaining 9 ``Easy'' videos contain minimal interactions (mean $\sim$1). We provide a standardized benchmark split consisting of 13 validation and 12 test videos.

\section{Detection and Tracking}
We utilize a \textsc{YOLOv8\textsubscript{Nano}}~\cite{Varghese2024YOLOv8AN} fine-tuned on our gorilla data to handle complex wildlife environments. Chosen for computational efficiency, it matches the performance of larger alternatives with a final mAP@50 of 95.4\%.

\subsection{Multi-Object Tracking Evaluation}
We evaluate three state-of-the-art multi-object trackers on \gorillatracking{}: ByteTrack~\cite{Zhang2021ByteTrackMT}, BoTSORT~\cite{Aharon2022BoTSORTRA}, and BoostTrack++~\cite{Stanojevic2024BoostTrackUT}. As shown in \autoref{tab:combined_tracking}, BoTSORT achieves the highest Higher Order Tracking Accuracy (HOTA) of 0.79, excelling in detection precision. However, for conservation workflows where maintaining a consistent identity is critical, BoostTrack++ shows better performance. It achieves the highest Identification F1 (IDF1) score of 0.75 and reduces Identity Switches (IDSW) by 37\% compared to BoTSORT. This stability comes at a computational cost: BoostTrack++ (110.6s) is $3.3\times$ slower than BoTSORT (33.1s).

We further investigated the impact of appearance models within the BoostTrack++ framework. Comparing a motion-only baseline against integrations with DINOv2 and our specialized Gorilla re-ID model reveals only marginal performance gains (reducing IDSW from 7.17 to 6.75). This plateau indicates that the primary tracking challenges in this domain, severe occlusions and merged detections during social interactions, render visual features ambiguous or unavailable. As a result, even domain-specific appearance embeddings provide limited benefit over robust motion modeling in these high-complexity scenarios.

\begin{table}[ht!]
    \centering
    \resizebox{0.9\linewidth}{!}{
        \begin{tabular}{l c c c}
            \toprule
            \textbf{Method} & \textbf{HOTA} $\uparrow$ & \textbf{IDF1} $\uparrow$ & \textbf{IDSW} $\downarrow$ \\
            \midrule
            ByteTrack \cite{Zhang2021ByteTrackMT} & 0.77 & 0.74 & 10.75 \\
            BoTSORT \cite{Aharon2022BoTSORTRA}    & \textbf{0.79} & 0.73 & 10.67 \\
            BoostTrack++               & 0.76 & \textbf{0.75} & \textbf{6.75} \\
            \midrule
            \multicolumn{4}{l}{\textit{BoostTrack++ re-ID Ablation Study}} \\
            w/o re-ID                & 0.76 & 0.75 & 7.17 \\
            w/ DinoV2-Giant                         & 0.75 & 0.74 & 6.92 \\
            w/ Gorilla re-ID (Ours)                 & 0.76 & 0.75 & \textbf{6.75} \\
            \bottomrule
        \end{tabular}
    }
    \caption{Tracking performance on the \gorillatracking{} dataset. The top section compares leading algorithms. The bottom section analyzes the impact of different appearance models within the BoostTrack++ framework. Other DINOv2 model sizes performed identically to DINOv2-Giant and are omitted.}
    \label{tab:combined_tracking}
    \vspace{-2mm}
\end{table}

\section{Re-Identification Framework}
The re-ID component forms the core of our pipeline, matching gorilla individuals across different video encounters. We evaluate solely on open-set re-identification using a strict \textbf{cross-encounter} protocol: for every query image (probe), we dynamically remove all images belonging to the same encounter from the gallery. This prevents the model from exploiting temporal proximity (e.g., adjacent videos), forcing reliance on robust biometric features. We use k-Nearest Neighbors (k-NN) ($k=5$) for retrieval and report Top-1 Accuracy, balanced over individuals.

For \gorillawild{}, the gallery comprises the \textit{respective evaluation split} (validation or test), the \textit{entire training set}, and the set of \textit{single-encounter individuals}. While single-encounter individuals cannot serve as probes, their inclusion acts as an additional source of ``distractor'' identities, expanding the search space to reflect real-world difficulty. For the \gorillazoo{} benchmark, the gallery consists exclusively of the zoo dataset.

\subsection{Frame-Level Embedding}

\begin{figure}[ht]
    \centering
    \includegraphics[width=\linewidth]{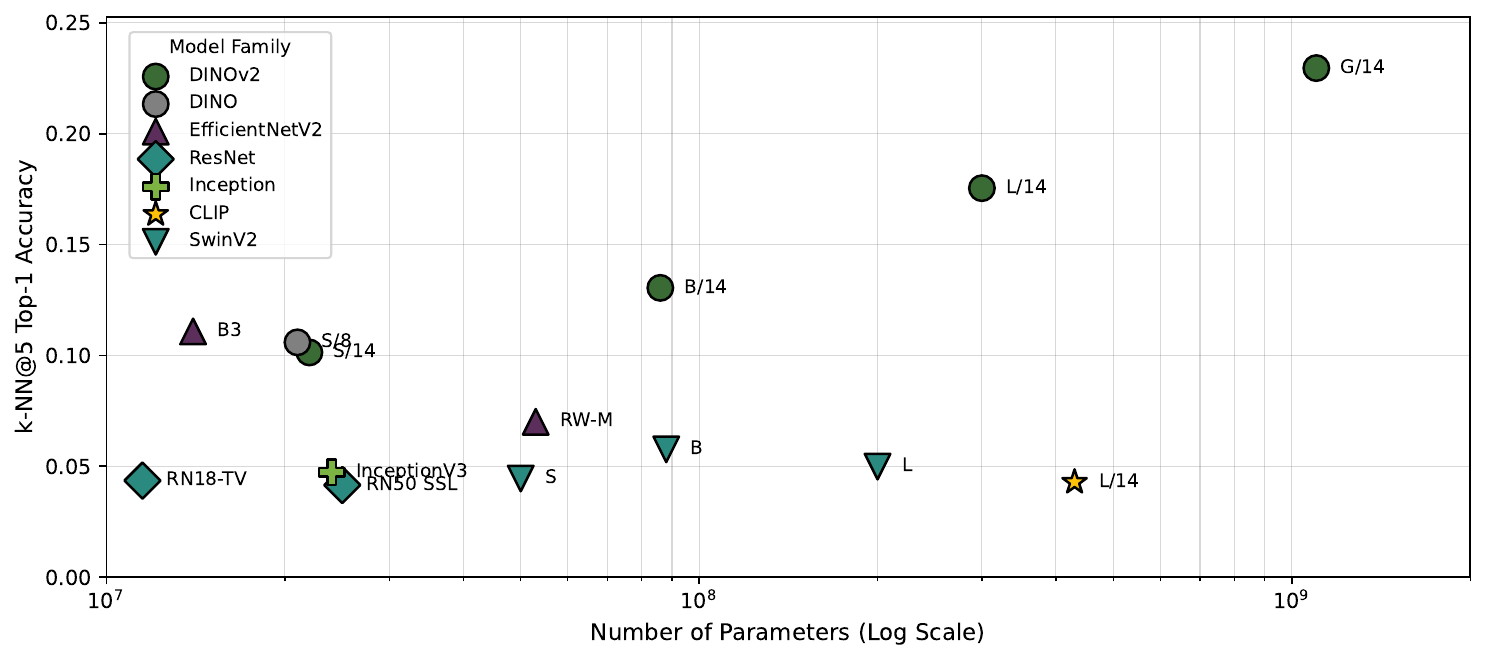}
    \caption{Zero-shot performance comparison of different pretrained models on the \gorillawild{} validation set.}
    \label{fig:pretrained_comparison}
    \vspace{-2mm}
\end{figure}

We evaluate various pretrained models on our validation data as shown in \autoref{fig:pretrained_comparison}. The comparison includes DINOv2~\cite{oquab_dinov2_2023}, CLIP~\cite{radford_learning_2021}, Swin Transformer~\cite{liu_swin_2022}, ResNet variants~\cite{he_deep_2015}, EfficientNet~\cite{tan_efficientnetv2_2021}, and Inception~\cite{szegedy_rethinking_2015}. DINOv2 variants achieve the best performance and are used for subsequent experiments.

\subsubsection{Fine-Tuning Image Models}
\label{sec:image-fine-tuning}
\begin{figure}[ht]
    \centering
    \includegraphics[width=\linewidth]{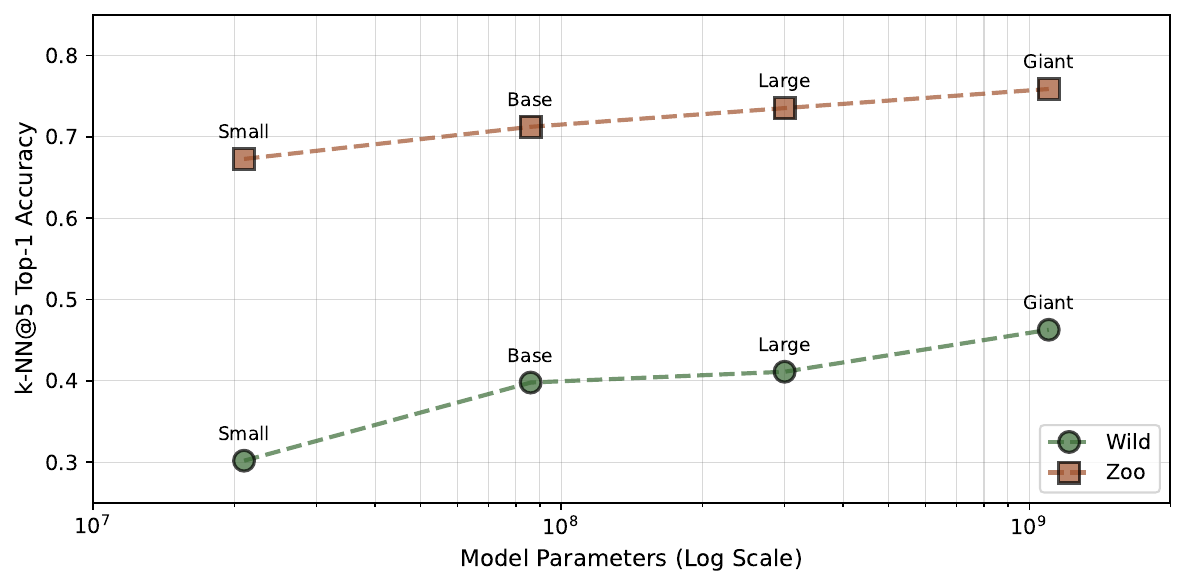}
    \caption{Top-1 accuracy of DINOv2 variants after supervised fine-tuning on the \gorillawild{} training set, evaluated on the \gorillawild{} and \gorillazoo{} test sets.}
    \label{fig:finetuned_comparison}
    \vspace{-2mm}
\end{figure}

We fine-tune models using hard triplet loss~\cite{schroff_facenet_2015}, applying a linear layer to reduce features to 256-dimensional embeddings from the class token output. Models are trained for 100 epochs with a batch size of 8 and accumulation over 6 steps. The maximum learning rate is $1.9 \times 10^{-7}$ using Adam and a cosine learning rate scheduler.  We set L2 and L2SP regularization ($\lambda_{L2} \approx 0.0059$, $\lambda_{L2SP} \approx 1.3 \times 10^{-5}$) to prevent overfitting. We train all models on a single Nvidia H100 GPU. 
We report the results in \autoref{fig:finetuned_comparison}.

\subsubsection{Temporal Self-Supervised Pretraining}
Annotating identities of individual gorillas is labor-intensive and requires domain expertise, limiting the scalability of supervised training. Prior work largely operates on single images \cite{caron_emerging_2021, he_masked_2021, chen_empirical_2021, oquab_dinov2_2023, iashin_self-supervised_2025}. We instead propose a multi-frame pretraining scheme that harnesses natural variation within video tracklets to improve training stability.

Our approach extends the standard DINOv2~\cite{oquab_dinov2_2023} multi-crop augmentation strategy by incorporating temporal diversity. Rather than sampling all crops from a single image, we aggregate crops across time, preserving identity while introducing natural pose and illumination changes. This yields a form of real-world augmentation that captures intra-individual variation more faithfully than synthetic transformations alone.

We evaluate three configurations: (i) a single-frame baseline matching the original DINOv2 implementation; (ii) a four-frame variant that samples every second frame from the first eight frames of a tracklet; and (iii) a ten-frame variant that uses the middle eight frames plus the first and last frames (as global crops). We sample 59,316 tracklets from the same original source as \gorillawild{}, while excluding its videos. For comparability, we retain the original DINOv2 hyperparameters for tuning.

\begin{figure*}[ht]
    \centering
    \includegraphics[width=0.6\linewidth]{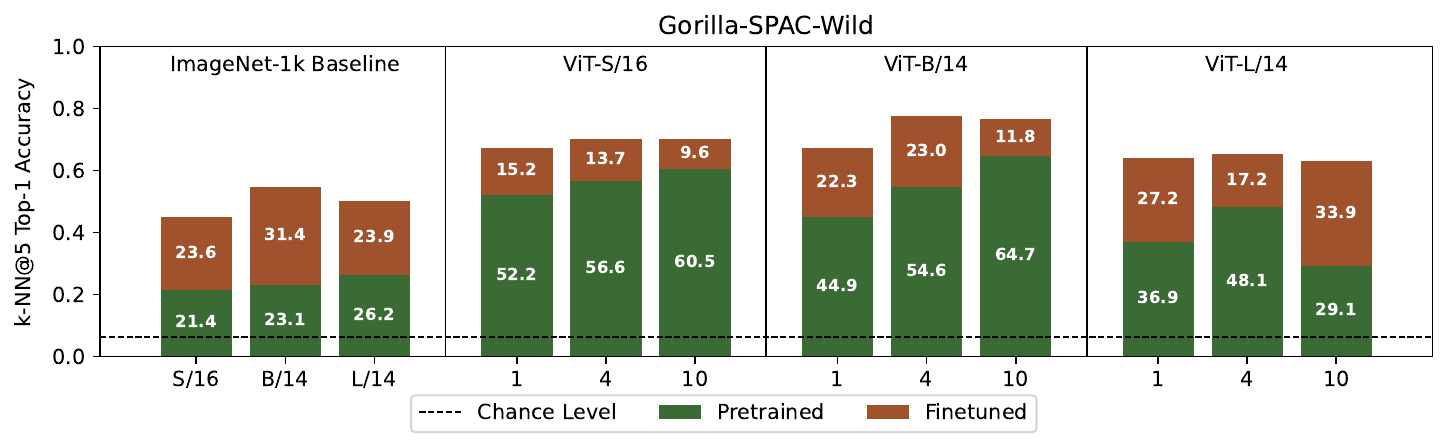}
\caption{Effect of Multi-frame Pretraining on the \gorillawild{} test set: We evaluate performance across single-frame, 4-frame, and 10-frame temporal sequences. The green bars denote accuracy immediately after pretraining, while the brown bars indicate the accuracy change ($\Delta$) after fine-tuning.}\label{fig:multi_frame_comparison}
\end{figure*}

The results in \autoref{fig:multi_frame_comparison} show that multi-frame pretraining benefits the small and base models, but not consistently the large model. For ViT-S/16 and ViT-B/14, both multi-frame variants outperform single-frame pretraining, before and after fine-tuning. The largest gain after pretraining is 19.8 percentage points for ViT-B/14 with ten frames (44.9\% to 64.7\%); after fine-tuning, the largest gain is 10.4 points for ViT-B/14 with four frames (67.2\% to 77.6\%). For ViT-L/14, four frames improve over single-frame pretraining (48.1\% vs.\ 36.9\% pretrained, 65.3\% vs.\ 64.1\% fine-tuned), whereas ten frames fall below it (29.1\% pretrained, 63.0\% fine-tuned). On \gorillazoo{} (not shown), ViT-L/14 accuracy after pretraining decreases with more frames (60.2\%, 53.3\%, and 48.3\% for 1, 4, and 10 frames), and single-frame pretraining also remains best after fine-tuning. Domain-specific pretraining on gorilla faces surpasses generic ImageNet-1k pretraining for all model sizes. Training time is comparable between single- and multi-frame setups. Notably, model scale alone does not determine performance: after fine-tuning, the domain-pretrained ViT-B/14 (76.5\% with ten frames) clearly outperforms the generically pretrained DINOv2-Giant fine-tuned on the same training set (46.3\%, \autoref{fig:finetuned_comparison}).

\subsubsection{Explainable Embeddings via Differentiable Proxies}
For specialized tasks like gorilla re-ID, where training data is scarce, ensuring that the model generalizes based on meaningful biological traits, such as nose prints and brow ridges, rather than spurious background correlations is critical. This necessity is underscored by recent findings from ~\citet{brookes_panaf-fgbg_2025}, which demonstrated that background information alone acts as a strong predictor in wildlife video analysis, leading to severe shortcut learning and poor out-of-distribution generalization. To verify that our model relies on discriminative biometric features rather than these environmental shortcuts, we employ AttnLRP~\cite{achtibat_attnlrp_2024} to generate faithful, pixel-level attribution maps.

AttnLRP is designed for classification models that produce continuous logits suitable for gradient-based propagation. However, our re-ID model outputs high-dimensional embeddings, where predictions rely on k-Nearest Neighbors, a non-differentiable operation. To bridge this gap, we propose and evaluate three differentiable proxy scores that approximate the k-NN decision while remaining faithful to the embedding geometry. Let $q$ be the query embedding, $\mathcal{P}$ the set of ``friend'' embeddings (same identity), and $\mathcal{N}$ the set of ``foe'' embeddings.

\begin{enumerate}
    \item \textbf{Similarity Score:} A direct measure of alignment that computes the cosine similarity $S_{\text{sim}} = q^T x_p$ between the query and a randomly sampled positive neighbor $x_p \in \mathcal{P}$.
    \item \textbf{Proto Margin Score:} Inspired by the hard triplet loss, this score calculates the margin between the query's similarity to a weighted ``friend prototype'' ($\mu_{\mathcal{P}}$) and a ``foe prototype'' ($\mu_{\mathcal{N}}$) constructed from the top-$k$ hardest negatives.
    \item \textbf{k-NN Margin Score:} To mimic the discrete voting process differentiably, we compute a softmax distribution over the database similarities. The score is defined as the margin between the total probability mass assigned to friends versus foes: $S_{\text{knn}} = \sum_{x_i \in \mathcal{P}} \sigma(q, x_i) - \sum_{x_j \in \mathcal{N}} \sigma(q, x_j)$.
\end{enumerate}
These proxies transform neighborhood relations into smooth scalar values, ensuring (i) continuity for stable propagation, (ii) faithfulness to embedding relations, and (iii) compatibility with automatic differentiation. 

\begin{figure}[ht]
    \centering
    \includegraphics[width=\linewidth]{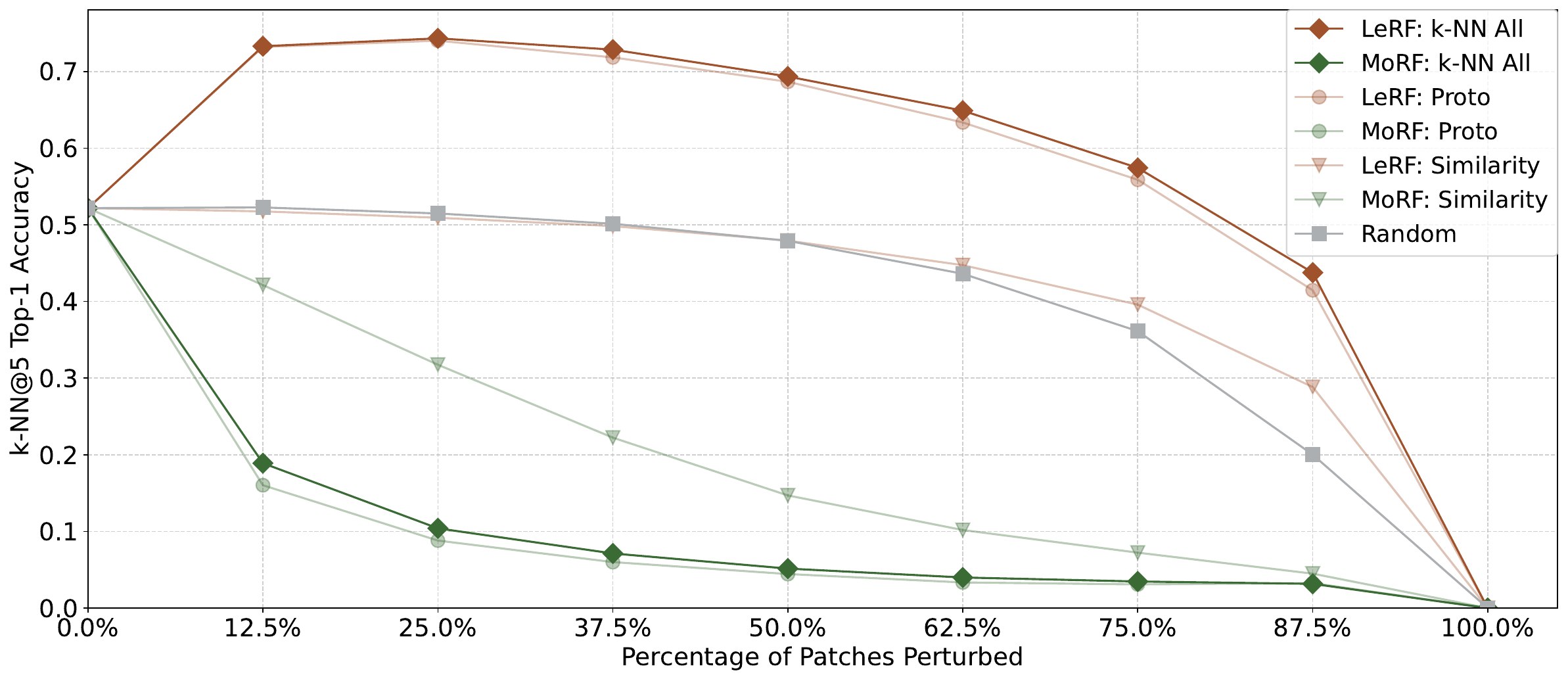}
    \caption{Faithfulness evaluation of our proposed AttnLRP adaptation using the patch flipping paradigm, validating the effectiveness of our proxy scores.}
    \label{fig:attnlrp_eval}
\end{figure}

\paragraph{Faithfulness Evaluation}
To quantify how accurately these explanations reflect the model's decision-making, we employ the patch flipping paradigm~\cite{samek_evaluating_2015}. We generate perturbation curves by ordering patches from Most Relevant First (MoRF) and Least Relevant First (LeRF). As shown in \autoref{fig:attnlrp_eval}, removing the most relevant patches causes a rapid drop in re-ID accuracy, whereas removing the least relevant patches preserves performance. Notably, over 90\% of the accuracy is maintained using only the top 25\% most relevant patches, confirming that the k-NN Margin Score provides the most faithful approximation of the model's dense retrieval process.

\begin{figure}[htbp]
    \centering
    \begin{subfigure}[t]{0.32\linewidth}
        \includegraphics[width=\linewidth]{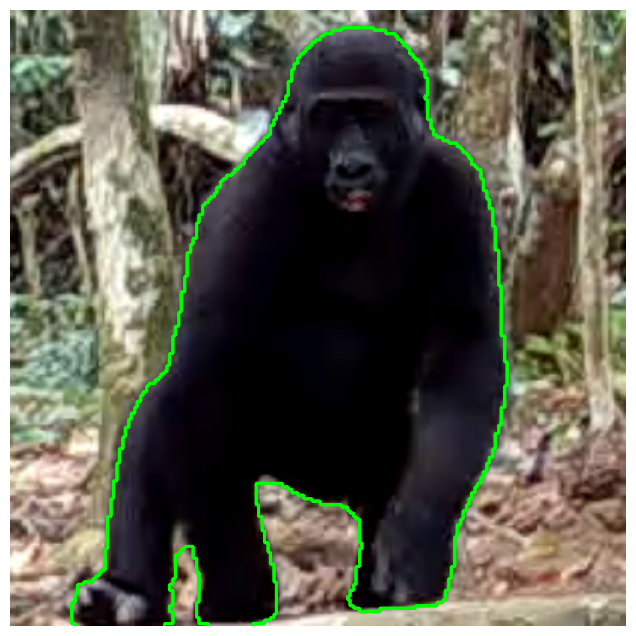}
    \end{subfigure}
    \hfill
    \begin{subfigure}[t]{0.32\linewidth}
        \centering
        \includegraphics[width=\linewidth]{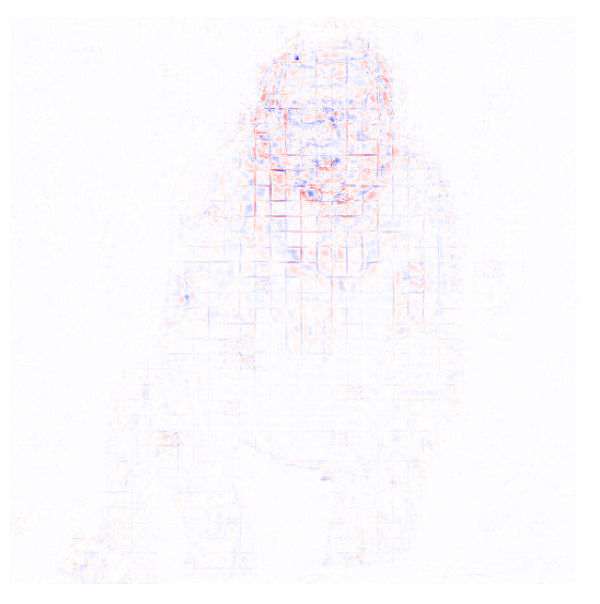}
    \end{subfigure}
    \hfill
    \begin{subfigure}[t]{0.32\linewidth}
        \centering
        \includegraphics[width=\linewidth]{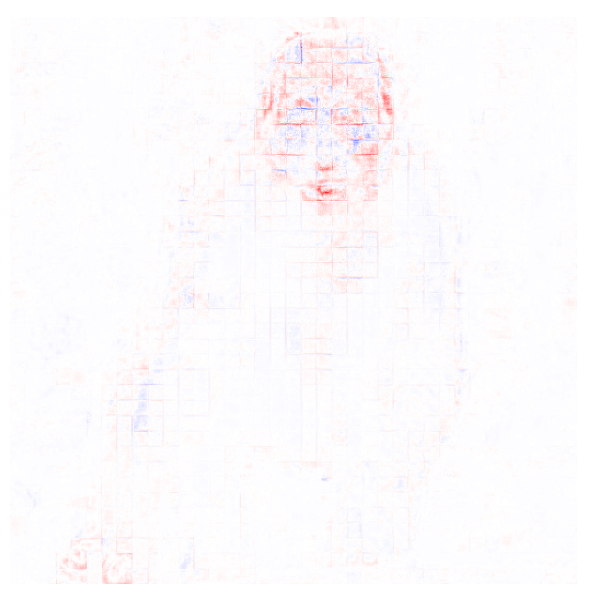}
    \end{subfigure}
    
    \caption{Comparison of relevance attributed to image pixels for original (center) and fine-tuned (right) DINOv2 models. Fine-tuning led to relevance being more concentrated on the face and body of the gorilla instead of being dispersed all over the image (including background).}
    
    \label{fig:attention_maps}
\end{figure}

We visually compare the relevance heatmaps of the original DINOv2 against our fine-tuned model in \autoref{fig:attention_maps}. The off-the-shelf DINOv2 exhibits dispersed attention, often focusing on background vegetation—aligning with the background reliance warnings raised by \citet{brookes_panaf-fgbg_2025}. In contrast, our fine-tuned model concentrates relevance on discriminative biometric features, specifically the face and body shape. Quantitatively, on the \gorillawild{} dataset, 92\% of the total relevance in our fine-tuned model falls within the ground-truth gorilla segmentation masks, compared to only 83\% for the baseline. Manual inspection reveals that the remaining 8\% of ``background'' relevance often highlights secondary gorillas present in the frame, indicating the model captures social context where appropriate.

\subsection{Tracklet-Level Feature Aggregation}
To produce a single decision for an entire tracklet, we evaluate several methods for aggregating sequences of frame-level feature vectors. Each frame-level feature is extracted using our fine-tuned \textit{DINOv2\textsubscript{Giant}} model from \autoref{sec:image-fine-tuning}. Given the stronger frame-level accuracy of the domain-pretrained ViT-B/14 (\autoref{fig:multi_frame_comparison}), combining multi-frame pretrained backbones with tracklet-level aggregation is a promising direction for future work.

We evaluate four aggregation strategies: \textit{majority voting}, where the final identity is determined by the most frequent top-1 prediction across all frames in a tracklet; \textit{confidence averaging}, where individual frame predictions are weighted by their retrieval confidence scores before aggregation; and \textit{embedding averaging}, which computes a single mean feature vector for the entire tracklet.

For a more sophisticated approach, we adapt the Aggregated Appearance Encoder (AAE) framework from \citet{zhu_sharc_2023}. This adaptation was necessary because our DINOv2-backbone produces global feature vectors for each frame, unlike the spatial feature maps used by the original network's Spatial-Temporal Aggregation Module (STAM)~\cite{wang_pyramid_2021}. Our primary modification is therefore to replace the STAM with a transformer-based mechanism called the \textit{Pairwise Transformer Aggregation Module} (PTAM). At each level of a pyramid structure, the PTAM takes two input embeddings from the level below, treats them as a two-item sequence, and processes them through a shallow transformer encoder for context-aware fusion. The resulting output is then fused through a linear projection back to the input dimensionality. We train the aggregation module using our frozen fine-tuned backbone. 

\subsection{Video-Based Architecture}
While ensemble methods aggregate static features, they fail to capture dynamic behavioral cues such as gait. To explicitly model these spatio-temporal features, we adapt several video models originally developed for action recognition to the task of representation learning for re-identification. We evaluate three distinct approaches: Adapter-based Image Model (\textbf{AIM})~\cite{yang_aim_2023}, which represents a parameter-efficient approach by freezing a pretrained ViT and injecting lightweight, trainable adapters to learn temporal relationships; \textbf{TimeSformer}~\cite{bertasius_is_2021-1} as an example of full spatio-temporal fine-tuning, where the model's self-attention mechanism is extended to jointly process information across space and time; and \textbf{InternVideo2}~\cite{wang_internvideo2_2024}, a state-of-the-art video foundation model, to gauge the effectiveness of general-purpose video representations on our specialized task.

To adapt these models for re-identification, we replace their final classification head with a linear projection layer to output a 256-dimensional embedding. We then fine-tune them using the same hard triplet loss objective function employed for our image-based models to ensure a fair comparison. Further, we investigate the impact of backbone initialization. Since standard implementations of AIM and TimeSformer initialize their backbones from supervised ViT checkpoints pretrained on ImageNet-1K, we additionally initialize them with the corresponding DINOv2\textsubscript{Base} checkpoint. This ensures that any observed performance differences result solely from the architectural approach rather than the quality of the underlying pretraining.

\subsection{Comparative Analysis}

\begin{table}[ht!]
\centering
\resizebox{\linewidth}{!}{
\begin{tabular}{llcc}
\toprule
\textbf{Method} & \textbf{Strategy / Backbone} & \textbf{\gorillawild{}} & \textbf{\gorillazoo{}} \\
\midrule
\multirow{4}{*}{Ensemble}   & Majority Voting       & 67.22 & 84.42 \\
                            & Confidence Averaging  & 69.91 & \textbf{84.75} \\
                            & Embedding Averaging   & 70.08 & 80.61 \\
                            & PTAM                  & \textbf{73.06} & 82.44 \\
\midrule
\multirow{2}{*}{AIM}        & ViT                   & 22.32 & 53.56 \\
                            & DINOv2-Base          & 7.02 & 34.50 \\
\midrule
\multirow{2}{*}{TimeSformer}& ViT                   & 25.75 & 64.59 \\
                            & DINOv2-Base          & 38.27 & 71.76 \\
\midrule
\multicolumn{2}{l}{InternVideo2}                    & 39.51 & 65.09 \\
\bottomrule
\end{tabular}
}
\caption{Re-identification performance (Top-1 Accuracy \%) on the \gorillawild{} test set and \gorillazoo{}.}
\label{tab:ensemble_video_eval}
\end{table}

The results of our comparative analysis, presented in Table~\ref{tab:ensemble_video_eval}, show a decisive finding: the temporal aggregation of robust static features significantly outperforms end-to-end video architectures for wildlife re-identification.

Our ensemble methods demonstrate superior performance, with PTAM achieving a top-1 accuracy of 73.06\% on \gorillawild{}, and simple \textit{Embedding Averaging} reaching 70.08\%. In stark contrast, dedicated video models struggle. InternVideo2 achieves only 39.51\%. We attribute this disparity primarily to the massive scale and extensive pretraining of DINOv2. While such large-scale foundation models are established for static images, developing a video-native equivalent of similar magnitude remains an open challenge; the prohibitive computational cost and complexity of pretraining on vast video archives have so far prevented the community from replicating this scale in the temporal domain.

To determine if we could bridge this gap by simply transferring these powerful features, we replaced the standard backbones of AIM and TimeSformer with DINOv2 checkpoints. However, even with this robust initialization, TimeSformer achieves only 38.27\%. This persistent failure suggests that simply ``modifying'' a foundation model optimized for static spatial features is insufficient. We hypothesize that these spatio-temporal architectures suffer from high hyperparameter sensitivity in data-scarce regimes, and that fully unlocking their potential likely requires a dedicated video-aware pretraining stage rather than direct adaptation of image weights.

\section{Clustering}
Our framework relies on clustering of embeddings for population size estimation, specifically, by clustering the features extracted from videos without identity labels and using the resulting number of clusters as an estimate of the total population. We evaluate several clustering algorithms, including Hierarchical Agglomerative Clustering (HAC)~\cite{Ward01031963}, 
DBSCAN~\cite{ester1996density}, and HDBSCAN~\cite{campello_density-based_2013}.

To improve accuracy, we use a constrained clustering approach that leverages the spatio-temporal metadata inherent in camera trap videos along with domain expertise. We enforce three specific constraints: (1) \textit{Frame difference} (cannot-link), ensuring individuals appearing in the same frame are treated as distinct; (2) \textit{Location difference} (cannot-link), separating individuals recorded simultaneously at different camera sites; and (3) \textit{Social group} (cannot-link), which utilizes expert annotations to prevent the merging of individuals belonging to different social groups. Constraint (3) requires expert knowledge of group membership; the approach therefore needs no identity labels, but it is not fully unsupervised.

We evaluate performance using Adjusted Rand Index (ARI)~\cite{rand1971objective} and Adjusted Mutual Information (AMI)~\cite{vinh2009information}, which measure the similarity between the resulting clusters and the ground-truth identities.

The results in Table~\ref{tab:cluster_results} show that unconstrained clustering severely over-segments \gorillawild{}, predicting 113--465 individuals for a true population of 16. On \gorillawild{}, integrating constraints improves ARI and AMI for all three algorithms and brings the population estimates of HAC and DBSCAN close to the ground truth ($K=17$ and $13$ respectively, vs.\ $16$). HDBSCAN, however, fragments further, growing from 113 to 204 clusters. On \gorillazoo{}, constraints do not reliably help: AMI improves for HAC and HDBSCAN, but ARI drops for HAC (0.201 to 0.180) and DBSCAN (0.038 to 0.011), whose AMI drops as well. All constrained cluster counts (44--249) remain far from the true $K=5$, and HDBSCAN moves from under-segmentation (2 clusters) to over-segmentation (249 clusters). The confined zoo environment generates significantly fewer distinct spatiotemporal conflicts; because individuals all belong to a single social group and constantly reside in a small, shared area, the constraints are far less effective at disambiguating identities than in the geographically dispersed wild setup. 

\begin{table}[ht!]
    \centering
    \resizebox{\linewidth}{!}{
    \begin{tabular}{cl ccc ccc}
        \toprule
         & & \multicolumn{3}{c}{\textbf{\gorillawild{}}} & \multicolumn{3}{c}{\textbf{\gorillazoo{}}} \\
        \cmidrule(lr){3-5} \cmidrule(lr){6-8}
        & \textbf{Method} & \textbf{ARI} $\uparrow$ & \textbf{AMI} $\uparrow$ & \textbf{\# C} & \textbf{ARI} $\uparrow$ & \textbf{AMI} $\uparrow$ & \textbf{\# Cluster} \\
        \midrule
        \multirow{3}{*}{\rotatebox[origin=c]{90}{\resizebox{3.6em}{!}{Unconstrained}}}
          & HAC      & 0.606 & 0.728 & 465 & 0.201 & 0.249 & 2277 \\
          & DBSCAN   & 0.379 & 0.625 & 341 & 0.038 & 0.116 & 596  \\
          & HDBSCAN  & 0.114 & 0.510 & 113 & 0.001 & 0.005 & 2    \\
        \midrule
        \multirow{3}{*}{\rotatebox[origin=c]{90}{\resizebox{3.6em}{!}{Constrained}}}
          & HAC      & 0.837 & 0.891 & 17  & 0.180 & 0.500 & 156  \\
          & DBSCAN   & 0.586 & 0.873 & 13  & 0.011 & 0.064 & 44   \\
          & HDBSCAN  & 0.184 & 0.677 & 204 & 0.048 & 0.465 & 249  \\
        \bottomrule
    \end{tabular}
    }
    \caption{Population counting performance evaluation. On \gorillawild{}, constraints improve ARI and AMI for all algorithms and bring HAC and DBSCAN close to the true population size; on \gorillazoo{}, the effect of constraints is mixed. The true population sizes are $K=16$ for \gorillawild{} and $K=5$ for \gorillazoo{}.}
    \label{tab:cluster_results}
\end{table}

\section{Conclusion}

In this work, we presented \textbf{GorillaWatch}, a unified framework designed to overcome the critical bottleneck in automated wildlife monitoring: the immense manual effort required to analyze vast video archives. By introducing \gorillawild{}, \gorillazoo{}, and \gorillatracking{}, we provide the research community with the first comprehensive benchmark suite for open-set primate re-identification and tracking, specifically curated to capture the challenging lighting, occlusion, and pose variations of the rainforest.

Our extensive technical evaluation offers several pivotal insights for applying deep learning to conservation. First, we provided a conclusive answer to the architectural debate between specialized video transformers and ensemble-based image models. Our results demonstrate that aggregating features from large-scale image foundation models (e.g., DINOv2) significantly outperforms end-to-end video architectures. We attribute this to the ``data scarcity gap'': While video models offer theoretical advantages for temporal modeling, they require massive training corpora to converge. In contrast, the robust, general-purpose representations from image foundation models prove far more transferable to the fine-grained task of individual identification.

Second, we demonstrated that multi-frame self-supervised pretraining is an effective way to bridge the gap between image models and video data. By modifying the standard training recipe to leverage temporal consistency within unlabelled tracklets, we achieved gains of up to 10.4 percentage points after fine-tuning over single-frame baselines for the small and base models, although the large model did not benefit consistently. This finding is particularly encouraging for conservation, as it suggests that the vast amounts of unlabelled archival footage currently sitting on hard drives can be actively used to improve model robustness without requiring expensive human annotation.

Finally, we addressed the ``black box'' problem inherent in deep learning through a novel, differentiable adaptation of AttnLRP. In wildlife monitoring, models frequently succumb to the ``Clever Hans'' effect, learning background scenery rather than animal identity. By visualizing the embedding decision process, we quantitatively verified that our fine-tuning strategy shifts the model's focus from environmental correlations (e.g., vegetation) to discriminative biometric traits, such as facial structure and body shape. This interpretability is essential for establishing scientific trust in automated census systems.

We publicly release all code, models, and datasets to serve as a blueprint for scalable, non-invasive monitoring.

\section*{Acknowledgments}
The project on which this report is based was funded by the Federal Ministry of Research, Technology and Space under the funding code “KI-Servicezentrum Berlin-Brandenburg” 16IS22092. We acknowledge the support of Sabine Plattner African Charities (SPAC) for their funding to this research. We are grateful to Zoo Berlin for their expert assistance and facility access. This collaboration enabled the development of AI tools capable of being deployed in the wild to directly support gorilla conservation. The responsibility for the content of this publication remains with the authors.

{
    \small
    \bibliographystyle{ieeenat_fullname}
    \bibliography{references_gorillawatch_updated}
}

\end{document}